\documentclass[letterpaper]{article} 
\usepackage{aaai2026}  
\usepackage{times}  
\usepackage{helvet}  
\usepackage{courier}  
\usepackage[hyphens]{url}  
\usepackage{graphicx} 
\usepackage{natbib}  
\usepackage{caption} 
\usepackage{algorithm}
\usepackage{algorithmic}
\usepackage{booktabs}
\usepackage{multirow}
\usepackage{siunitx}
\usepackage{amssymb}
\usepackage{amsmath}
\usepackage{newfloat}
\usepackage{listings}
\DeclareCaptionStyle{ruled}{labelfont=normalfont,labelsep=colon,strut=off} 
\floatstyle{ruled}
\newfloat{listing}{tb}{lst}{}
\floatname{listing}{Listing}
\title{UniMo: Unified Motion Generation and Understanding with Chain of Thought}

\author{
    Guocun Wang\textsuperscript{\rm 1}\equalcontrib, 
    Kenkun Liu\textsuperscript{\rm 2}\equalcontrib, 
    Jing Lin\textsuperscript{\rm 3}, 
    Guorui Song\textsuperscript{\rm 1}, 
    Jian Li\textsuperscript{\rm 1}\thanks{Corresponding author.}, 
    Xiaoguang Han\textsuperscript{\rm 2,4,5}\footnotemark[2]
}
\affiliations{
    \textsuperscript{\rm 1}SIGS, Tsinghua University\\
    \textsuperscript{\rm 2}SSE, The Chinese University of Hong Kong (Shenzhen)\\
    \textsuperscript{\rm 3}MMLab, Nanyang Technological University\\
    \textsuperscript{\rm 4}FNii-Shenzhen\\
    \textsuperscript{\rm 5}Guangdong Provincial Key Laboratory of Future Networks of Intelligence\\
    \{wang-gc25, sgr24, l-j21\}@mails.tsinghua.edu.cn, kenkunliu@link.cuhk.edu.cn, jing026@e.ntu.edu.sg, hanxiaoguang@cuhk.edu.cn
}

\begin{document}

\maketitle

\begin{abstract}
Existing 3D human motion generation and understanding methods often exhibit limited interpretability, restricting effective mutual enhancement between these inherently related tasks. While current unified frameworks based on large language models (LLMs) leverage linguistic priors, they frequently encounter challenges in semantic alignment and task coherence. Moreover, the next-token prediction paradigm in LLMs is ill-suited for motion sequences, causing cumulative prediction errors. To address these limitations, we propose UniMo, a novel framework that integrates motion-language information and interpretable chain of thought (CoT) reasoning into the LLM via supervised fine-tuning (SFT). We further introduce reinforcement learning with Group Relative Policy Optimization (GRPO) as a post-training strategy that optimizes over groups of tokens to enforce structural correctness and semantic alignment, mitigating cumulative errors in motion token prediction. Extensive experiments demonstrate that UniMo significantly outperforms existing unified and task-specific models, achieving state-of-the-art performance in both motion generation and understanding.
\end{abstract}

\begin{links}
    \link{Code}{https://github.com/GuocunWang/UniMo}
\end{links}

\begin{figure*}[ht]
    \centering
    \includegraphics[width=0.98\textwidth]{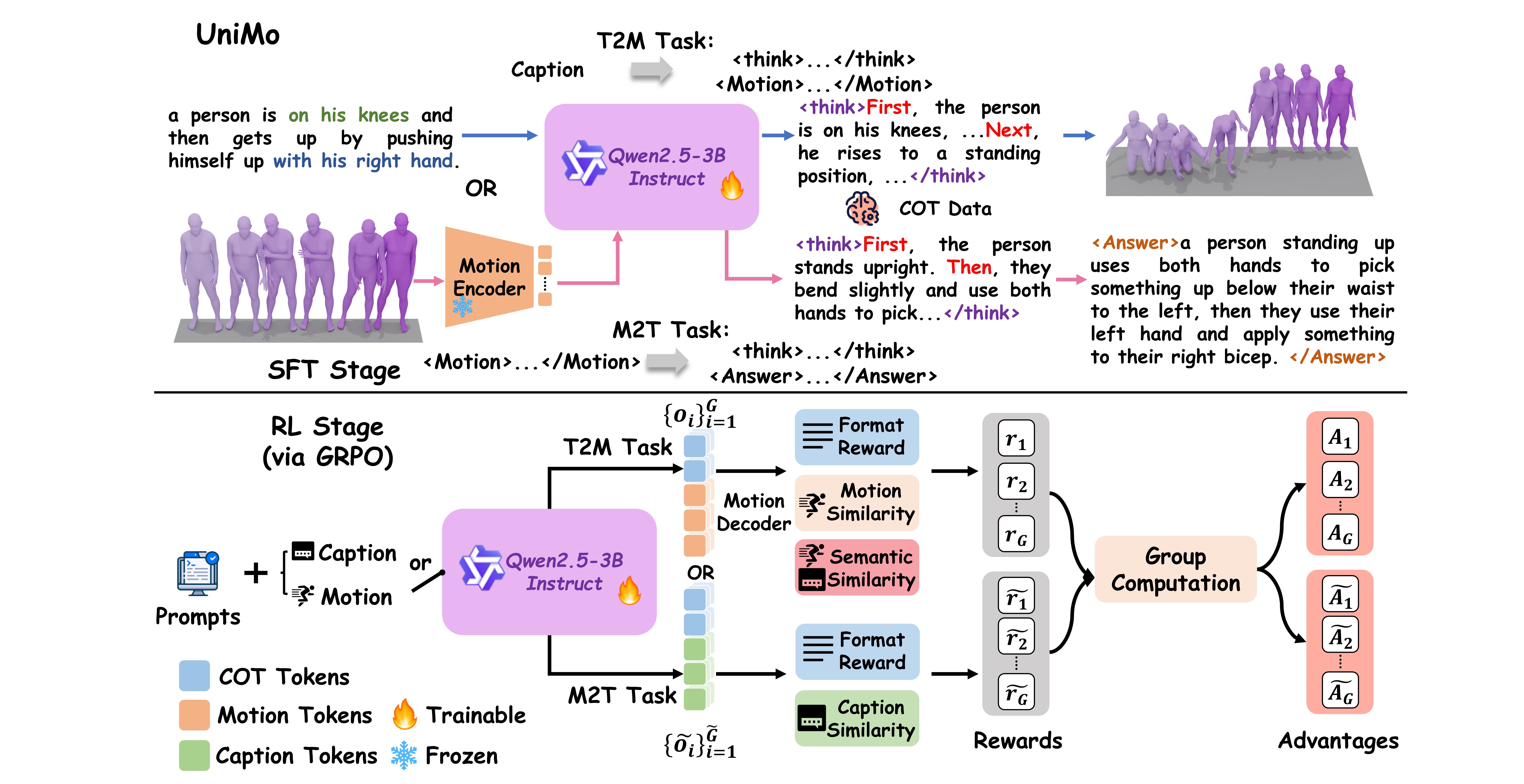} 
    \caption{The UniMo is trained in two stages: the SFT stage and the reinforcement learning stage with GRPO. In the SFT stage, the model is teached to perform both T2M and M2T tasks with structured reasoning, i.e. CoT. In the RL stage, the model is further optimized with task-specific rewards, enabling unified and interpretable motion generation and understanding.}
    \label{fig:UniMo}
\end{figure*}

\section{Introduction}

Human motion generation and understanding is a pair of inverse problems. The formal aims to synthesize human motion from given input like textual instructions, and is a popular research field due to its various applications in animation, controllable video generation, gaming, virtual reality and etc \cite{hu2024animate,zhu2024champ,qiu2025lhm,wang2024disco,song2025towards}. By contrast, human motion understanding is to interpret the actions and intentions in the form of text descriptions from the given human motion. Previous methods \cite{MLD,MotionDiffuse,MotionR1,Momask} usually treat them as two tasks and address them separately. Despite their state-of-the-art performance in each single task, the mutually-inverse property of the two tasks is overlooked and these methods merely translate one modality to the other. Some efforts \cite{MotionLLM,MotionGPT,Motiongpt2} attempt to unify the two tasks into a single LLM-based model, but their performance in each single task is still unsatisfactory.

There are several reasons accounting for the situation. The first is that the given textual instruction is usually quite simple and coarse, so there might be many human motions could follow the instruction. In other words, the correspondence between the text and motion is too weak. 
Therefore, existing methods can hardly learn to generate fine-grained human actions. Second, unlike natural language that has strong sequential causality, human motion is essentially temporal sequence with weaker causality. Thus, the next-token-prediction paradigm is not the optimal training manner.

Therefore, to address the above mentioned issues, we propose the UniMo, the first unified model for motion generation and understanding with chain of thought (CoT). The model is built upon a pretrained open-sourced LLM, i.e. Qwen2.5-3B-Instruct \cite{qwen2_5}, to exploit their compressed knowledge. The key ingredients of of our model includes motion-consistent CoT data curation, CoT-guided motion generation and understanding, and RL-based joint training with task-specific rewards. 

In the aspect of CoT data, MotionR1 \cite{MotionR1}, as a concurrent work, exploits the existing LLM and merely extend the original texts from the motion-language dataset like HumanML3D \cite{T2M} into step-by-step CoT data. Although simple and format-aligned, they cannot ensure the constructed CoT data is consistent with the corresponding motions. Using this data may lead to severe hallucination for trained model. By contrast, we render the motions via 3D Engine like Blender into videos, and then exploit the existing VLM model like Qwen2.5-VL \cite{qwen2_5_vl} to describe the human motion in the rendered videos. The output descriptions are used as the CoT data, which are consistent with the original human motions.

With motion-consistent CoT data, we exploit them for CoT-guided unified modeling. For the task of text-to-motion (T2M) generation, instead of directly predicting motion tokens with input text, our model requires to first plan the motion generation process in the format of step-by-step CoT, then start to generate motion tokens conditioned on the input text and the CoT. Similarly, for the task of the motion-to-text (M2T), the model also need to output the CoT before generate the final text description. By introducing the CoT as a bridge into our unified model, it provides more precise control signal for T2M generation and decompose the complex problem of M2T into easier subproblems.

As mentioned before, next-token-prediction paradigm is not optimal for the motion token prediction, so we resort to RL for post-training. By employing the Group Relative Policy Optimization (GRPO), we can force the model to follow the desired format and design rewards to increase the performance of the two tasks, which optimize multiple tokens simultaneously instead of a single token in next token prediction. Moreover, we jointly train the two tasks together and finally find that they can benefit each other.

With these designs, UniMo finally achieves state-of-the-art performance on both tasks, demonstrating the synergic effect of unifying generation and understanding with CoT for the first time. In all, our contributions are as follows:
\begin{itemize}
    \item We propose UniMo, an LLM-based framework that unifies both motion generation and understanding with chain of thought reasoning.
    \item We design task-specific rewards for GRPO to supervise the RL-based joint training of both T2M and M2T tasks, which optimizes over groups of tokens and mitigates cumulative errors from next-token prediction.
    \item We curate a CoT dataset based on HumanML3D, which bridges the gap between motion and language.
    \item Our unified model outperforms existing state-of-the-art single-task methods in both tasks, proving that unified modeling can simultaneously improve motion generation and understanding through bidirectional synergy. 
\end{itemize}

\section{Related Works}
\subsection{Human Motion Generation}
Earlier works \cite{language2pose,ghosh2021synthesis,huang2020dance,plappert2018learning} adopt a deterministic paradigm and usually obtain unsatisfactory performance. Then, the stochastic methods \cite{T2M,TM2T,MLD,MotionDiffuse,MDM,Momask} gradually dominate the field and the text input also becomes a main kind of condition due to its flexibility. Representative GAN-based methods \cite{cai2018deep,wang2020learning} are action-conditioned. With increasing popularity of diffusion models \cite{ho2020denoising,nichol2021improved,song2020denoising}, some pioneer works like MDM \cite{MDM}, MotionDiffuse \cite{MotionDiffuse}, MLD \cite{MLD} successfully apply diffusion models to the text-to-motion generation task and achieve competitive performance. Besides them, MoMask \cite{Momask} adopts mask modeling for motion tokens and achieves the state-of-the-art performance. There are also some attempts \cite{T2M-GPT,MotionGPT,Motiongpt2,MotionChain,MotionR1} adopting the auto-regressive model. However, these AR-based models are still inferior to other methods in performance.

\subsection{Unified Modeling of Motion Generation and Understanding}
There are already some multi-modal LLMs \cite{xie2024show,chen2025janus,wang2024emu3,tong2024metamorph,pan2025transfer,deng2025emerging,chen2025blip3} that attempt to unify the visual understanding and generation in a single model, like GPT-4o, EMU-3 \cite{wang2024emu3} and Bagel \cite{deng2025emerging}. They excel in diffusion-based models \cite{esser2024scaling,lipman2022flow} in the aspects of complex instruction following and multiturn editing. Some previous works \cite{Motiongpt2,MotionLLM,MotionGPT,zhang2024motiongpt} also explore the unified modeling of human motion generation and understanding. However, their performance is inferior to that of single-task models, due to the straggle between the two tasks.

\subsection{Chain of Thought for Generative Models}
Chain of thought (CoT) has been proven to be the key to unlocking the reasoning ability of LLMs \cite{wei2022chain,Deepseek-r1} as CoT can decompose a complex problem into multiple subproblems that are easier to solve. Drawing on the similar idea, some works propose to synthesize images with CoT \cite{deng2025emerging,guo2025can}, i.e., generate detailed generation steps first, and then generate the image according to these detailed descriptions. With CoT, these methods could generate images from indirect prompts, requiring further thinking. MotionR1 \cite{MotionR1}, as a concurrent work, is the first to adopt CoT to enhance human motion generation. However, their CoT data is curated via a rewrite manner, which cannot ensure the consistency with original human motion.

\begin{figure}[ht]
    \centering
    \includegraphics[width=\linewidth]{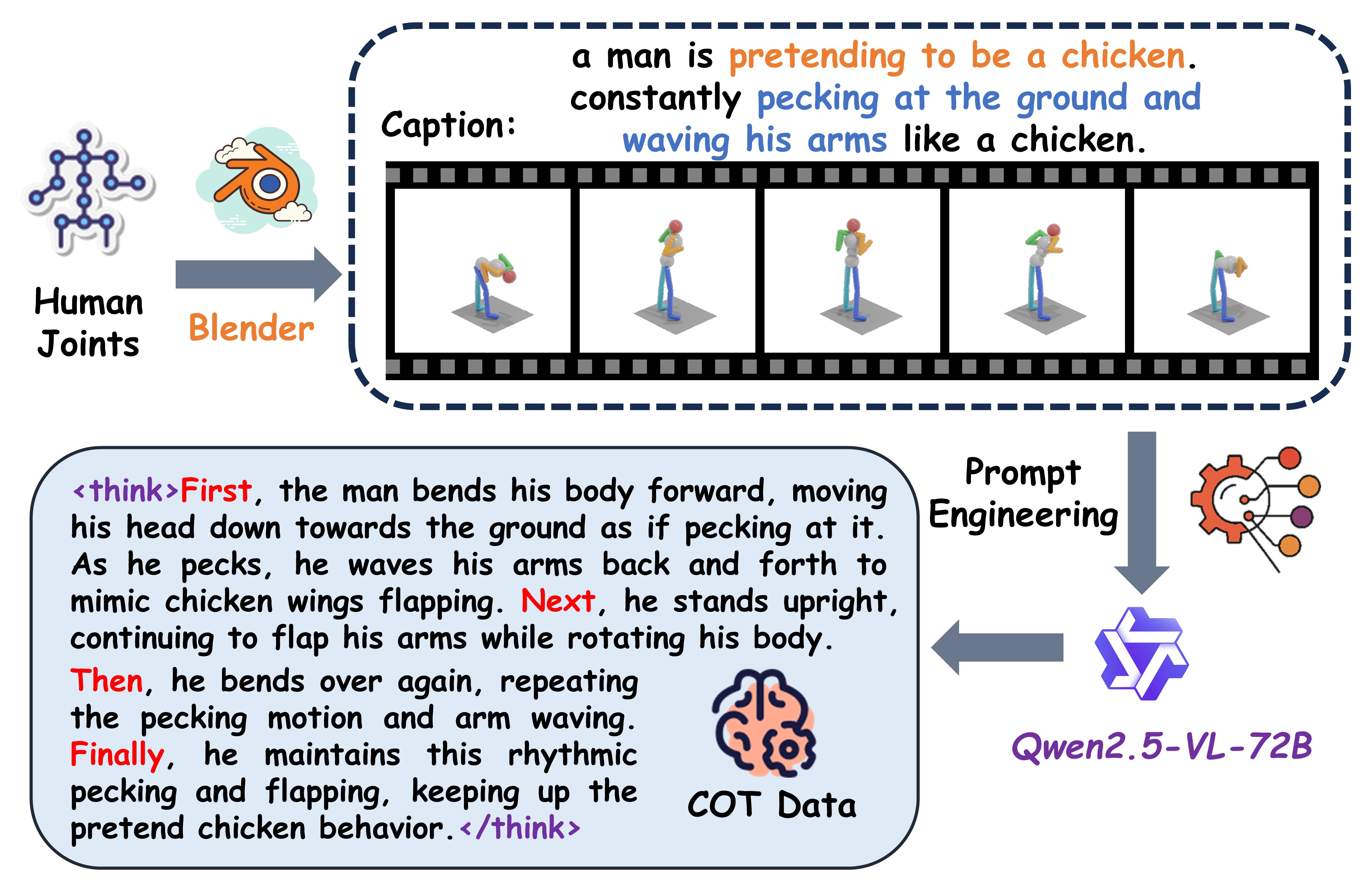}
    \caption{Illustration of the CoT annotation process. Human joint sequences are rendered in the Blender and paired with captions, which are further processed by the Qwen2.5-VL-72B to generate reasoning traces.}
    \label{fig:COT-Annotation}
\end{figure}

\begin{figure*}[!htbp]
    \centering
    \includegraphics[width=0.96\textwidth]{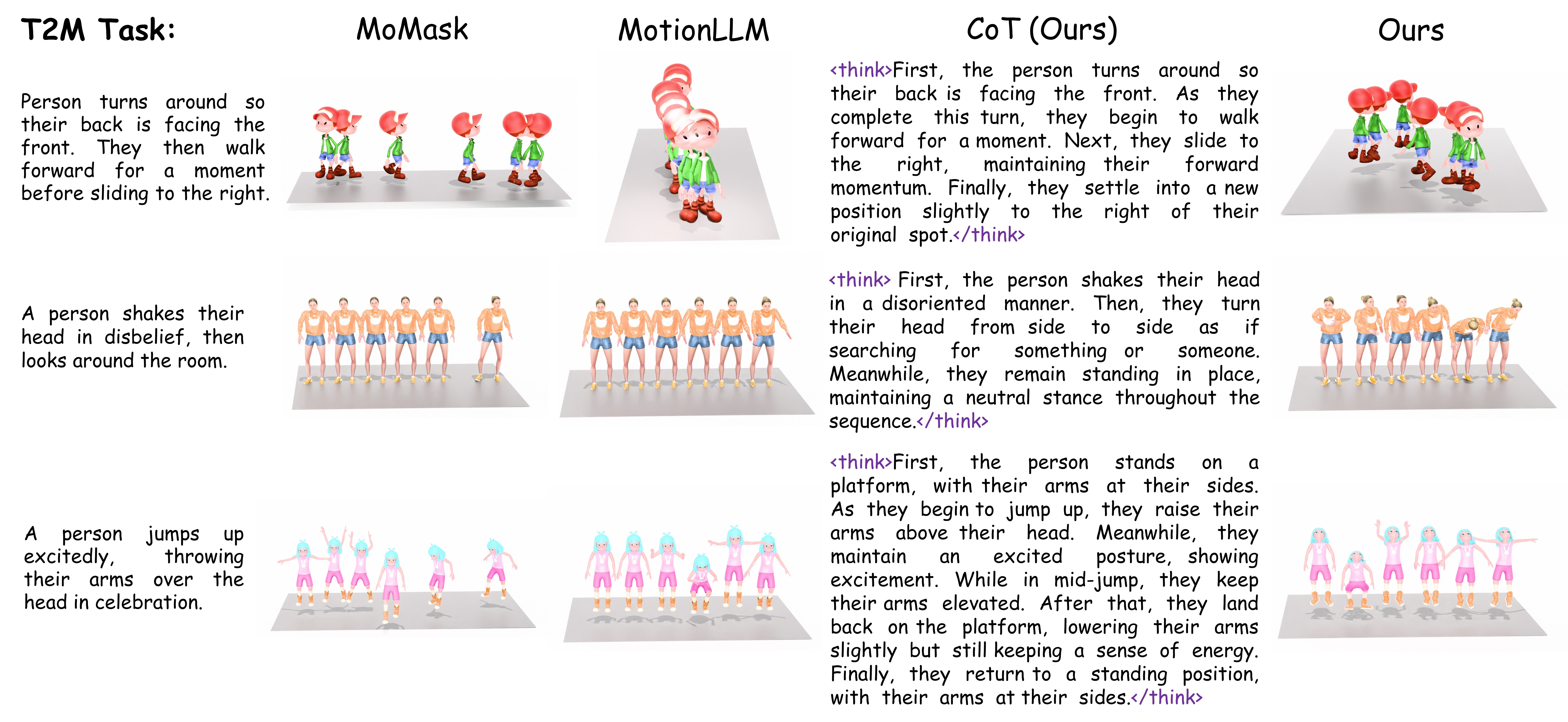}
    \caption{Qualitative comparison of our method with other open-source SOTAs such as MoMask \cite{Momask} and MotionLLM \cite{MotionLLM} for the text-to-motion task. Our method presents stronger instruction-following capability and can generate sequential actions. We highly recommend the readers to watch the video comparisons in supplementary materials.}
    \label{fig:t2m}
\end{figure*}

\begin{figure*}[ht]
    \centering
    \includegraphics[width=\textwidth]{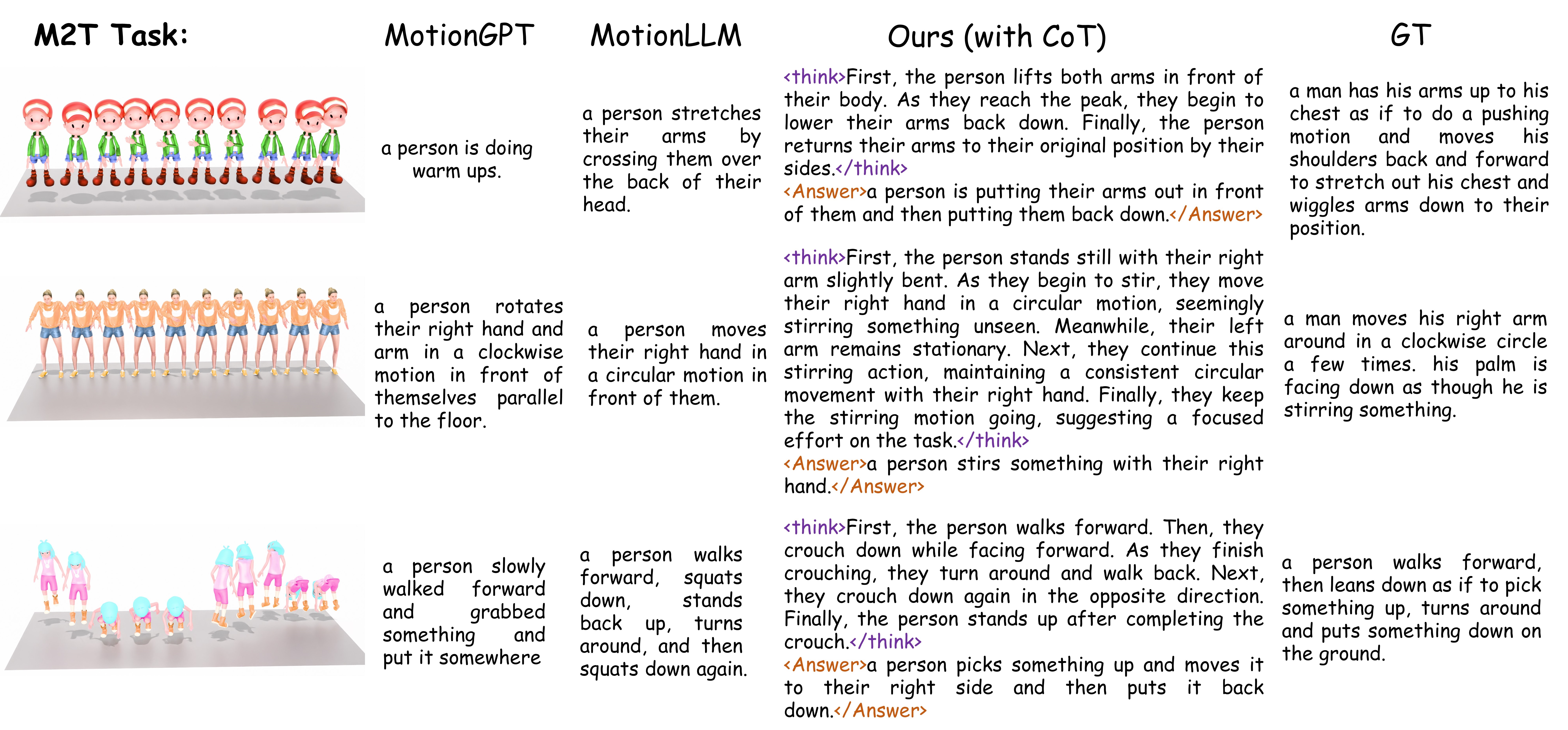}
    \caption{Qualitative comparison of our method with other open-source SOTAs such as MotionGPT \cite{MotionGPT} and MotionLLM \cite{MotionLLM} for the motion-to-text task. Our method can more precisely describe complex motions.}
    \label{fig:m2t}
\end{figure*}

\section{Method}

\subsection{Unified Modeling}

As illustrated in Figure~\ref{fig:UniMo}, UniMo incorporates structured, interpretable reasoning through CoT and achieves tight coupling between motion generation and understanding by joint training across tasks. The model is trained in two stages: the SFT stage with CoT annotations as a cold start, and a reinforcement learning stage using GRPO \cite{GRPO} with task-specific rewards.

\subsection{Motion Representation}

To integrate human motion into the LLM framework, we adopt the VQ-VAE as the motion tokenizer, following prior work \cite{T2M-GPT, MotionGPT, MotionLLM}. This module discretizes continuous motion sequences into token indices, so they can be processed in the same manner as the LLM's tokens. 
Formally, given a human motion sequence expressed as a time series $\mathbf{m}_{1:T}\in\mathbb{R}^{T\times D}$, where $T$ is the number of frames and $D$ is the dimension of each frame. The encoder $E$ maps this input to a latent feature sequence $\mathbf{z}_{1:T'} \in R^{T' \times d}, \quad T' = T/l$, where \(d\) is the latent dimension and \(l\) is the downsampling factor. Each latent vector $\mathbf{z}_t$ is then quantized with a learnable codebook $\mathbf{C} = \{\mathbf{c}_k\}_{k=1}^{K}$, where $K$ is the number of discrete motion codes and each $\mathbf{c}_k \in \mathbb{R}^d$. The quantized representation can be represented as:
\begin{eqnarray}
\hat{\mathbf{z}}_t = \arg\min_{\mathbf{c}_k \in \mathcal{C}} \|\mathbf{z}_t - \mathbf{c}_k\|_2.
\end{eqnarray}

To reconstruct the original motion sequence, a decoder $D$ maps the quantized tokens back into the continuous space: $\hat{\mathbf{m}}_{1:T} = D(\hat{\mathbf{z}}_{1:T'})$. The motion tokenizer is trained using a composite loss that includes the reconstruction loss $\mathcal{L}_{\text{recon}}$, the embedding loss $\mathcal{L}_{\text{embed}}$ and the commitment loss $\mathcal{L}_{\text{commit}}$. Following the prior work \cite{T2M-GPT}, the objective function is:
\begin{eqnarray}
\mathcal{L}_{\text{VQ}}& =&
\mathcal{L}_{\text{recon}}
+ \underbrace{\left\| \mathbf{z}_{1:T'} - \text{sg}[\hat{\mathbf{z}}_{1:T'}] \right\|_2}_{\mathcal{L}_{\text{embed}}}\\
&& + \underbrace{\left\| \text{sg}[\mathbf{z}_{1:T'}] - \hat{\mathbf{z}}_{1:T'} \right\|_2}_{\mathcal{L}_{\text{commit}}} \nonumber ,
\end{eqnarray}
where $\text{sg}[\cdot]$ denotes the stop-gradient operator. We employ exponential moving average updates for the codebook, along with a codebook reset strategy. The VQ-VAE is frozen during subsequent training stages.

\subsection{Supervised Fine-tuning with CoT}

In the SFT stage, UniMo leverages high-quality CoT annotations to bridge the gap between natural language and human motion. As shown in the Figure~\ref{fig:COT-Annotation}, we first render the motion sequences in the HumanML3D dataset into video clips using Blender, and use the Qwen2.5-VL-72B \cite{qwen2_5_vl} to generate CoT annotations. For each motion-caption pair, the Qwen2.5-VL-72B is prompted with both the rendered video and its original caption as input, and then instructed to produce structured step-by-step reasoning under the format:\texttt{<think>\{CoT\}</think>}. Unlike Motion-R1 \cite{MotionR1}, which generates CoT solely from the captions using DeepSeek-R1 \cite{Deepseek-r1}, this multimodal annotation strategy ensures that the generated CoT is not only aligned with the caption semantics but also grounded in the physical realization of the motion, as observed from the video.

The introduction of curated CoT in the SFT stage offers two advantages over previous methods \cite{MotionR1,MotionLLM,MotionGPT}. First, training on paired motion, caption and CoT data encourages the model to generate reasoning aligned with real motion dynamics, thus enabling it to acquire robust semantic grounding and explicit interpretability within a unified framework. Second, this stage serves as a principled initialization, equipping the model with strong bidirectional mapping and structured reasoning abilities before reinforcement learning.

\subsection{Reinforcement Learning with GRPO}

\paragraph{Group Relative Policy Optimization.}
To further align the model outputs with structural and semantic expectations under the paradigm of reinforcement learning, we adopt GRPO \cite{GRPO}, a variant of PPO \cite{PPO} tailored for generation tasks. GRPO evaluates multiple sampled completions per input prompt, estimating the advantage function via within-group comparisons, which reduces variance in policy updates and stabilizes the learning process without requiring an explicit critic network. It integrates a clipping range with KL regularization, ensuring policy updates remain stable and consistent with the SFT-trained reference policy. The training objective is formally defined as:
\begin{eqnarray}
\mathcal{J}_{\text{GRPO}}(\theta) &=& E_c \left[ \frac{1}{G} \sum_{i=1}^{G} \min \left( 
\frac{\pi_{\theta}(o_i|q)}{\pi_{\text{old}}(o_i|q)} \hat{A}_i, \right. \right. \nonumber \\
&& \left. \left. \text{clip} \left( 
\frac{\pi_{\theta}(o_i|q)}{\pi_{\text{old}}(o_i|q)}, 1-\varepsilon, 1+\varepsilon \right) \hat{A}_i \right) \right] \nonumber \\
&& - \beta \cdot D_{\text{KL}}(\pi_{\theta} \parallel \pi_{\text{ref}}),
\end{eqnarray}
\begin{eqnarray}
\hat{A}_i = \frac{r_i - \text{mean}(\{r_1,...,r_G\})}{\text{std}(\{r_1,...,r_G\})},
\end{eqnarray}
where $q$ denotes the input prompt, and $o_i$ represents the $i$-th sampled output among a group of $G$ completions. The normalized advantage $\hat{A}_i$ is computed from the corresponding reward signals. The ratio $\frac{\pi_\theta(o_i|q)}{\pi_{\text{old}}(o_i|q)}$ measures the relative probability assigned to the same output $o_i$ under the new policy $\pi_\theta$ versus the old policy $\pi_{\text{old}}$. The clipping parameter $\varepsilon$ prevents drastic changes in the policy by restricting this ratio within a conservative range. Additionally, the KL divergence is used to regularize deviations from the reference policy $\pi_{\text{ref}}$, weighted by the coefficient $\beta$.

\paragraph{Format Reward.}
To ensure format compliance, we introduce the format reward $r_{format}$ that encourages the model to think before answering. For the T2M task, the required format is \texttt{<think>\{CoT\}</think><Motion>\{Motion Tokens\}</Motion>}. For the M2T task, it is \texttt{<think>} \texttt{\{CoT\}</think><Answer>\{Caption\}</Answer>}. Answers that strictly follow these formats get the reward of 1, otherwise they get 0.

\paragraph{Motion Similarity Reward.}
The motion similarity reward $r_{motion}$ is introduced to encourage realistic and coherent motion generation in the T2M task. To encourage the model to produce motions that closely resemble the true temporal dynamics and spatial patterns of the targets, it leverages the pretrained motion encoder $f_{\text{motion}}$ from \cite{T2M} to compute the cosine similarity between the generated motion $\mathbf{\hat{m}}$ and the ground-truth motion $\mathbf{m}$:
\begin{eqnarray}
r_{\text{motion}} = \frac{f_{\text{motion}}(\hat{\mathbf{m}}) \cdot f_{\text{motion}}(\mathbf{m})}{\|f_{\text{motion}}(\hat{\mathbf{m}})\|_2 \cdot \|f_{\text{motion}}(\mathbf{m})\|_2}.
\end{eqnarray}

\paragraph{Semantic Similarity Reward.}
To maintain semantic alignment between the input caption $T$ and the generated motion $\mathbf{\hat{m}}$ in the T2M task, the semantic similarity reward $r_{semantic}$ uses pretrained encoders $f_{\text{motion}}$ and $f_{\text{text}}$ to compute cosine similarity in a shared embedding space, which are from the prior work \cite{T2M}. This ensures generated motions faithfully reflect the input textual semantics. The formula for the semantic similarity reward is as follows:
\begin{eqnarray}
r_{\text{semantic}} = \frac{f_{\text{motion}}(\hat{\mathbf{m}}) \cdot f_{\text{text}}(T)}{\|f_{\text{motion}}(\hat{\mathbf{m}})\|_2 \cdot \|f_{\text{text}}(T)\|_2}.
\end{eqnarray}
\paragraph{Caption Similarity Reward.}
In the M2T task, the caption similarity reward $r_{caption}$ is obtained with the CLIP \cite{CLIP} text encoder $f_{\text{CLIP}}^{\text{text}}$ as the cosine similarity between the generated caption $\hat{T}$ and the reference caption $T$. This reward is then doubled to match the combined scale of the $r_{motion}$ and $r_{semantic}$ employed in the T2M task. This process is demonstrated in the following formula:
\begin{eqnarray}
r_{\text{caption}} = 2 \cdot \frac{f_{\text{CLIP}}^{\text{text}}(\hat{T}) \cdot f_{\text{CLIP}}^{\text{text}}(T)}{\|f_{\text{CLIP}}^{\text{text}}(\hat{T})\|_2 \cdot \|f_{\text{CLIP}}^{\text{text}}(T)\|_2}.
\end{eqnarray}
\begin{table*}[ht]
  \centering
  \footnotesize
  \setlength{\tabcolsep}{1mm}
  \begin{tabular}{cccccccc}
    \toprule
    \multirow{2}*{\textbf{Methods}} & \multicolumn{3}{c}{\textbf{R-Precision} $\uparrow$} & \multirow{2}*{\textbf{FID} $\downarrow$}  & \multirow{2}*{\textbf{MM-Dist}$\downarrow$} & \multirow{2}*{\textbf{Diversity}$\uparrow$} & \multirow{2}*{\textbf{MModality}$\uparrow$} \\
    \cline{2-4}
    & \textbf{Top1} & \textbf{Top2}& \textbf{Top3} \\
    \midrule
    MDM \cite{MDM} & $0.320^{\pm 0.005}$ & $0.498^{\pm 0.004}$ & $0.611^{\pm 0.007}$ & $0.544^{\pm 0.044}$ & $5.566^{\pm 0.027}$ & $9.559^{\pm 0.086}$ & \textbf{2.799}$^{\pm 0.074}$  \\
    MLD \cite{MLD} & $0.481^{\pm 0.003}$ & $0.673^{\pm 0.003}$ & $0.772^{\pm 0.002}$ & $0.473^{\pm 0.013}$ & $3.196^{\pm 0.010}$ & $9.724^{\pm 0.082}$ & $2.413^{\pm 0.072}$ \\
    MotionDiffuse \cite{MotionDiffuse} & $0.491^{\pm 0.001}$ & $0.681^{\pm 0.001}$ & $0.782^{\pm 0.001}$ & $0.630^{\pm 0.001}$ & $3.113^{\pm 0.001}$ & $9.410^{\pm 0.049}$ & $1.553^{\pm 0.064}$ \\
    \midrule
    T2M \cite{T2M}& $0.457^{\pm 0.002}$ & $0.559^{\pm 0.007}$ & $0.740^{\pm 0.003}$ & $1.067^{\pm 0.002}$ & $3.340^{\pm 0.008}$ & $9.188^{\pm 0.002}$ & $2.090^{\pm 0.088}$ \\
    TM2T \cite{TM2T} & $0.424^{\pm 0.003}$ & $0.618^{\pm 0.003}$ & $0.729^{\pm 0.002}$ & $1.501^{\pm 0.017}$ & $3.467^{\pm 0.011}$ & $8.589^{\pm 0.076}$ & \underline{2.424}$^{\pm 0.079}$ \\
    T2M-GPT \cite{T2M-GPT} & $0.491^{\pm 0.003}$ & $0.680^{\pm 0.003}$ & $0.775^{\pm 0.002}$ & \underline{0.116}$^{\pm 0.004}$ & $3.118^{\pm 0.011}$ & $9.761^{\pm 0.081}$ & $1.856^{\pm 0.111}$ \\
    MotionGPT \cite{MotionGPT} & $0.492^{\pm 0.003}$ & $0.681^{\pm 0.003}$ & $0.778^{\pm 0.002}$ & $0.232^{\pm 0.008}$ & $3.096^{\pm 0.008}$ & $9.528^{\pm 0.071}$ & $2.008^{\pm 0.083}$ \\
    MoMask \cite{Momask} & \underline{0.521}$^{\pm 0.002}$ & $0.713^{\pm 0.002}$ & $0.807^{\pm 0.002}$ & \textbf{0.045}$^{\pm 0.002}$ & $2.958^{\pm 0.008}$ & $9.620^{\pm 0.064}$ & $1.241^{\pm 0.064}$ \\
    MotionChain \cite{MotionChain} & 0.504$^{\pm 0.003}$ & 0.617$^{\pm 0.002}$ & 0.790$^{\pm 0.003}$ & 0.248$^{\pm 0.009}$ & 3.033$^{\pm 0.010}$ & 9.470$^{\pm 0.075}$ & 1.727$^{\pm 0.014}$ \\
    MotionLLM \cite{MotionLLM} & 0.515$^{\pm 0.004}$ & 0.691$^{\pm 0.003}$ & 0.801$^{\pm 0.004}$ & 0.230$^{\pm 0.009}$ & 2.967$^{\pm 0.020}$ & 9.908$^{\pm 0.102}$ & 2.142$^{\pm 0.014}$ \\
    MotionGPT-2 \cite{Motiongpt2}& 0.496$^{\pm 0.002}$ & 0.691$^{\pm 0.003}$ & 0.782$^{\pm 0.004}$ & 0.191$^{\pm 0.004}$ & 3.080$^{\pm 0.013}$ & 9.860$^{\pm 0.026}$ & 2.137$^{\pm 0.022}$ \\
    Motion-R1 \cite{MotionR1} & 0.515$^{\pm 0.003}$ & \underline{0.719}$^{\pm 0.002}$ & \underline{0.818}$^{\pm 0.002}$ & 0.201$^{\pm 0.004}$ & \underline{2.854}$^{\pm 0.010}$ & \underline{10.026}$^{\pm 0.075}$ & 2.317$^{\pm 0.105}$ \\
    \midrule
    UniMo (Ours) & \textbf{0.539}$^{\pm 0.003}$ & \textbf{0.738}$^{\pm 0.002}$ & \textbf{0.831}$^{\pm 0.002}$ & 0.177$^{\pm 0.004}$ & \textbf{2.768}$^{\pm 0.010}$ & \textbf{10.042}$^{\pm 0.076}$ & 1.924$^{\pm 0.080}$\\
    \bottomrule
  \end{tabular}
  \caption{Quantitative results of the T2M task on the HumanML3D dataset. Each evaluation is repeated 20 times with average metrics and 95\% confidence intervals. The best scores are highlighted in bold, and the second-best scores are underlined.}
  \label{tab:t2m_compare}
\end{table*}

\begin{table*}[!htbp]
\centering
\begin{tabular}{cccccc}
\toprule
\textbf{Captioning} & \textbf{BLEU@1↑} & \textbf{BLEU@4↑} & \textbf{ROUGE-L↑} & \textbf{CIDEr↑} & \textbf{BertScore↑} \\
\midrule
TM2T \cite{TM2T} & 48.90 &  8.27 & 38.1 & 15.80 & 32.2 \\
LaMPM2T \cite{Lamp} & 47.8 & 13.04 & 37.1 & 28.9 & 32.7 \\
MoTe \cite{MoTe} & 46.7 & 11.15 & 37.4 & 31.5 & 30.3 \\
MotionGPT \cite{MotionGPT} & 48.20 & 12.47 & 37.4 & 29.20 & 32.4 \\
MotionGPT-2 \cite{Motiongpt2} & 48.7 & 13.8 & 37.6 & 29.8 & 32.6 \\
MotionChain \cite{MotionChain} & 48.10 & 12.56 & 33.9 & 33.70 & 36.9 \\
MotionLLM \cite{MotionLLM} & \underline{54.53} & \underline{17.65} & \underline{48.7} & \underline{33.74} & \underline{42.63} \\
\midrule
UniMo (Ours) & \textbf{63.10} & \textbf{19.74} & \textbf{48.8} & \textbf{46.69} & \textbf{54.26} \\
\bottomrule
\end{tabular}
\caption{Quantitative results of the M2T task on the HumanML3D dataset. The best scores are highlighted in bold, and the second-best scores are underlined.}
\label{tab:m2t_compare}
\end{table*}

\begin{table*}[!htbp]
\centering
\footnotesize
\begin{tabular}{cccccccccccccc}
\toprule
\multirow{2}*{\textbf{CoT}} & \multirow{2}*{\textbf{$r_{\text{motion}}$}} & \multirow{2}*{\textbf{$r_{\text{semantic}}$}} & \multirow{2}*{\textbf{$r_{\text{caption}}$}}
& \multirow{2}*{\textbf{FID}$\downarrow$} & \multirow{2}*{\textbf{Diversity}$\uparrow$}
& \multicolumn{3}{c}{\textbf{R-Precision}↑} 
& \multirow{2}*{\textbf{B@1}$\uparrow$} & \multirow{2}*{\textbf{B@4}$\uparrow$}& \multirow{2}*{\textbf{R-L}$\uparrow$} & \multirow{2}*{\textbf{CIDEr}$\uparrow$} & \multirow{2}*{\textbf{Bert}$\uparrow$} \\
\cmidrule(lr){7-9}
& & & & & & \textbf{Top1} & \textbf{Top2} & \textbf{Top3} & & & & & \\
\midrule
 &  &  &  & \underline{0.178} & 9.873 & 0.384 & 0.559 & 0.659  & 56.64 & 16.34 & 44.2 & 36.56 & 49.01 \\
 & $ \checkmark $ &  & $ \checkmark $ & 0.218 & 9.694 & 0.509 & 0.706 & 0.807 & 61.91 & 19.15 & 47.5 & 45.26 & 52.94 \\
 &  & $ \checkmark $ & $ \checkmark $ & 0.255 & 9.937 & 0.534 & \textbf{0.738} & \textbf{0.834} & 60.99 & 18.86 & 47.3 & 44.52 & 52.68 \\
 & $ \checkmark $ & $ \checkmark $ & $ \checkmark $ & 0.180 & 9.771 & 0.533 & 0.735 & \underline{0.832} & 58.43 & 16.63 & 45.7 & 41.08 & 51.58 \\
 $ \checkmark $ &  &  &  & 0.292 & 9.726 & 0.460 & 0.642 & 0.735 & 55.34 & 15.31 & 43.2 & 32.08 & 48.16 \\
 $ \checkmark $ & $ \checkmark $ &  & $ \checkmark $ & 0.179 & \underline{9.994} & 0.518 & 0.712 & 0.808 & 62.75 & \underline{19.95} & \underline{48.4} & 44.45 & 53.98 \\
 $ \checkmark $ &  & $ \checkmark $ & $ \checkmark $ & 0.240 & 9.822 & \underline{0.537} & \underline{0.737} & 0.829 & \textbf{63.40} & \textbf{20.41} & \textbf{48.8} & \underline{45.94} & \textbf{54.42} \\
 $ \checkmark $ & $ \checkmark $ & $ \checkmark $ & $ \checkmark $ & \textbf{0.177} & \textbf{10.042} & \textbf{0.539} & \textbf{0.738} & 0.831 & \underline{63.10} & 19.74 & \textbf{48.8} & \textbf{46.69} & \underline{54.26} \\
\bottomrule
\end{tabular}
\caption{Ablation study on the effectiveness of CoT and different GRPO reward components on the HumanML3D dataset.}
\label{tab:ablation_cot_reward}
\end{table*}

\begin{table*}[ht]
\centering
\begin{tabular}{llcccccc}
\toprule
\textbf{Train Task} & \textbf{Stage} & \multicolumn{3}{c}{\textbf{R-Precision}↑} & \textbf{FID}↓ & \textbf{Diversity}↑ & \textbf{MM-Dist}↓ \\
\cmidrule(lr){3-5}
 & & \textbf{Top1} & \textbf{Top2} & \textbf{Top3} & & &\\
\midrule
T2M         & SFT          & 0.438 & 0.613 & 0.702 & \underline{0.201} & 9.748 & 3.588 \\
T2M         & SFT+RL     & \underline{0.529} & \textbf{0.739} & \textbf{0.832} & 0.203 & \underline{9.780} & \textbf{2.743} \\
T2M+M2T    & SFT          & 0.460 & 0.642 & 0.735 & 0.292 & 9.726 & 3.360\\
T2M+M2T    & SFT+RL     & \textbf{0.539} & \underline{0.738} & \underline{0.831} & \textbf{0.177} & \textbf{10.042} & \underline{2.768} \\
\bottomrule
\end{tabular}
\caption{Ablation study of the synergy effect of unified modeling for the T2M task.}
\label{tab:ablation_t2m_unified_modeling}
\end{table*}

\begin{table*}[!htbp]
\centering
\begin{tabular}{llccccc}
\toprule
\textbf{Train Task} & \textbf{Stage} & \textbf{BLEU@1}↑ & \textbf{BLEU@4}↑ & \textbf{ROUGE-L}↑ & \textbf{CIDEr}↑ & \textbf{BertScore}↑ \\
\midrule
M2T         & SFT          & 54.62 & 14.74 & 43.3 & 31.05 & 47.88 \\
M2T         & SFT+RL     & \underline{61.91} & \underline{18.89} & \underline{47.5} & \underline{42.13} & \underline{52.88} \\
T2M+M2T    & SFT          & 55.34 & 15.31 & 43.2 & 32.08 & 48.16 \\
T2M+M2T    & SFT+RL     & \textbf{63.10} & \textbf{19.74} & \textbf{48.8} & \textbf{46.69} & \textbf{54.26} \\
\bottomrule
\end{tabular}
\caption{Ablation study of the synergy effect of unified modeling for the M2T task.}
\label{tab:ablation_m2t_unified_modeling}
\end{table*}

\section{Experiments}

\subsection{HumanML3D with Curated CoT Annotations}

The HumanML3D dataset \cite{T2M} is a large-scale benchmark designed for evaluating human motion generation and understanding from natural language. It contains over 14,616 human motion clips sourced from AMASS \cite{AMASS} and HumanAct12 \cite{HumanAct12}, paired with nearly 44,970 textual descriptions. Each motion sequence is represented as a series of 3D joint positions over time, providing rich temporal and spatial information for modeling complex human behaviors. In this work, we further augment HumanML3D with curated CoT annotations, providing step-by-step reasoning traces that explicitly link language and motion.

\subsection{Evaluation Metrics}
For the T2M task, evaluation metrics from the \cite{T2M} include R-Precision at Top-1, Top-2, and Top-3, FID, MM-Dist, Diversity and MModality, which together reflect the realism and diversity of the generated motions. For the motion-to-text task, we report BLEU \cite{Bleu}, ROUGE-L \cite{Rouge}, CIDEr \cite{Cider} and BertScore \cite{Bertscore}, providing a comprehensive evaluation of the linguistic and semantic quality of the generated captions.

\subsection{Implementation Details}

UniMo is trained in two stages on $8 \times$A100 GPUs with the global batch size of 8, with Qwen2.5-3B-Instruct \cite{qwen2_5} serving as the foundational model. During the SFT stage, the model is first adapted to the motion modality by extending the vocabulary with motion tokens and pre-trained on the T2M task for 10 epochs. This is followed by joint fine-tuning on both T2M and M2T tasks for another 10 epochs. The Adam optimizer is employed with the learning rate of $1 \times 10^{-4}$, following the cosine decay.

In the RL stage, we employ the commonly-used GRPO \cite{GRPO}, and the training process is conducted for 14,000 steps, with the learning rate of $5 \times 10^{-5}$. Each prompt is completed with $G=8$ samples to compute the rewards, and gradient clipping with a maximum norm of 0.1 is applied. For stable policy optimization, GRPO utilizes the clipping range of $\varepsilon=0.2$, and the KL coefficient of $\beta=0.001$.

\subsection{Comparisons with the State of the Arts}

\paragraph{Text-to-Motion Results.} 

For the T2M task, we conduct comparisons with a comprehensive set of strong baselines, including diffusion-based methods, transformer-based methods, as well as recent LLM-based approaches. All evaluations follow the same experimental settings as previous studies to ensure fair and consistent comparison, with each method evaluated 20 independent runs and results reported as the average within a 95\% confidence interval.

The quantitative results for the motion generation task are presented in Table~\ref{tab:t2m_compare}. UniMo outperforms state-of-the-art methods in most metrics with an obvious margin. Although MoMask \cite{Momask} reports a lower FID, it sacrifices the MModality score at 1.241, whereas UniMo achieves a higher MModality score of 1.924. Overall, UniMo demonstrates a robust balance between motion realism and semantic expressiveness. Qualitative comparisons are shown in Figure \ref{fig:t2m}, our method exhibits stronger instruction-following ability and generates more coherent sequential actions than the open-source methods such as MoMask \cite{Momask} and MotionLLM \cite{MotionLLM}.

\paragraph{Motion-to-Text Results.}

Table~\ref{tab:m2t_compare} shows the results for the motion captioning task. UniMo achieves the new state-of-the-art results across all evaluation metrics. Compared to the strongest previous baseline MotionLLM \cite{MotionLLM}, UniMo shows the substantial improvement in all metrics. The results demonstrate the effectiveness of UniMo in generating motion captions that are more precise, relevant, and consistent with the input motion sequences. Figure \ref{fig:m2t} demonstrates the superior capability of our method in generating detailed and semantically aligned motion captions compared to other SOTA methods, i.e. MotionGPT \cite{MotionGPT} and MotionLLM \cite{MotionLLM}.

\subsection{Ablation Studies}

\paragraph{The effectiveness of CoT.}
Experimental results in Table~\ref{tab:ablation_cot_reward} demonstrate that introducing CoT leads to a significant improvement across multiple metrics. In the T2M task, CoT improves semantic alignment: Top-1 R-Precision increases from 0.384 to 0.460. Although FID slightly increases when CoT is used alone, combining CoT with the rewards achieves the lowest FID of 0.177 and the best balance across all metrics, confirming the overall benefit from the introduction of CoT. In the T2M task, BLEU@1 rises from 58.43 without CoT to 63.10 when CoT is applied, and CIDEr improves from 41.08 to 46.69. Similar gains are observed across BLEU@4, ROUGE-L and BertScore, demonstrating that CoT helps the model generate more semantically grounded and accurate motion descriptions.

\paragraph{The effectiveness of RL with task-specific rewards}
We further study the impact of reinforcement learning with task-specific rewards.
As shown in Table \ref{tab:ablation_cot_reward}, adding the motion similarity reward $r_{\text{motion}}$ notably reduces the FID from 0.240 to 0.177 when combined with CoT and other rewards. The semantic similarity reward $r_{\text{semantic}}$ has a marked impact on R-Precision, particularly boosting Top-1 to 0.539 when used together with CoT, $r_{\text{motion}}$ and $r_{\text{caption}}$. The caption similarity reward $r_{\text{caption}}$ significantly enhances the quality of generated descriptions in the M2T task, as shown by improvements in all textual metrics.

\paragraph{The synergy effect of unified modeling}

To investigate the effectiveness of unified modeling, we compare the performance of models trained solely on single tasks (T2M or M2T) with those trained jointly on both tasks. Experimental results are shown in Table~\ref{tab:ablation_t2m_unified_modeling} and ~\ref{tab:ablation_m2t_unified_modeling}. For the task of T2M, without RL, the unified modeling performs better than training T2M solely in all metrics. With RL, the unified modeling is also generally better than the single task of T2M. 
For the task of M2T, unified modeling is signficantly better than training the single task of M2T in all settings and all metrics. These results demonstrate the benefits of integrating generation and understanding tasks within a unified framework.

\section{Conclusion}

In this paper, we present UniMo, a unified framework for 3D human motion generation and understanding based on large language models. Our approach integrates CoT reasoning with GRPO to achieve joint modeling of T2M and M2T tasks in a structured and interpretable manner. Building upon the HumanML3D dataset, we construct a new motion-language dataset with motion-consistent CoT annotations. By designing task-specific rewards, UniMo achieves stronger semantic alignment between motion and language, and supports interpretable reasoning in both tasks. Importantly, we adopt RL with task-specific rewards, which enables the model to optimize over groups of tokens and mitigates the cumulative errors in the next-token prediction paradigm of LLMs for the motion token prediction. Extensive experiments demonstrate that UniMo achieves state-of-the-art results on both motion generation and understanding.

\bigskip
\section*{Acknowledgments}
This work was supported by the Shenzhen Science and Technology Project under Grant KJZD20240903103210014, in part by NSFC with Grant No. 62293482, the Basic Research Project No. HZQB-KCZYZ-2021067 of Hetao Shenzhen-HK S\&T Cooperation Zone, Guangdong Provincial Outstanding Youth Fund(No. 2023B1515020055), by Shenzhen Science and Technology Program No. JCYJ20220530143604010, NSFC No.62172348.

\bibliography{aaai2026}

\appendix

\twocolumn[
\begin{center}
{\LARGE\bfseries ---Supplementary Materials---\\[4ex]
UniMo: Unified Motion Generation and Understanding with Chain of Thought}
\end{center}
\vspace{12ex}
]

\section{Details about CoT Data Curation}

\subsection{Prompt for CoT Annotations}

\begin{figure*}[!htbp]
    \centering
    \includegraphics[width=\textwidth]{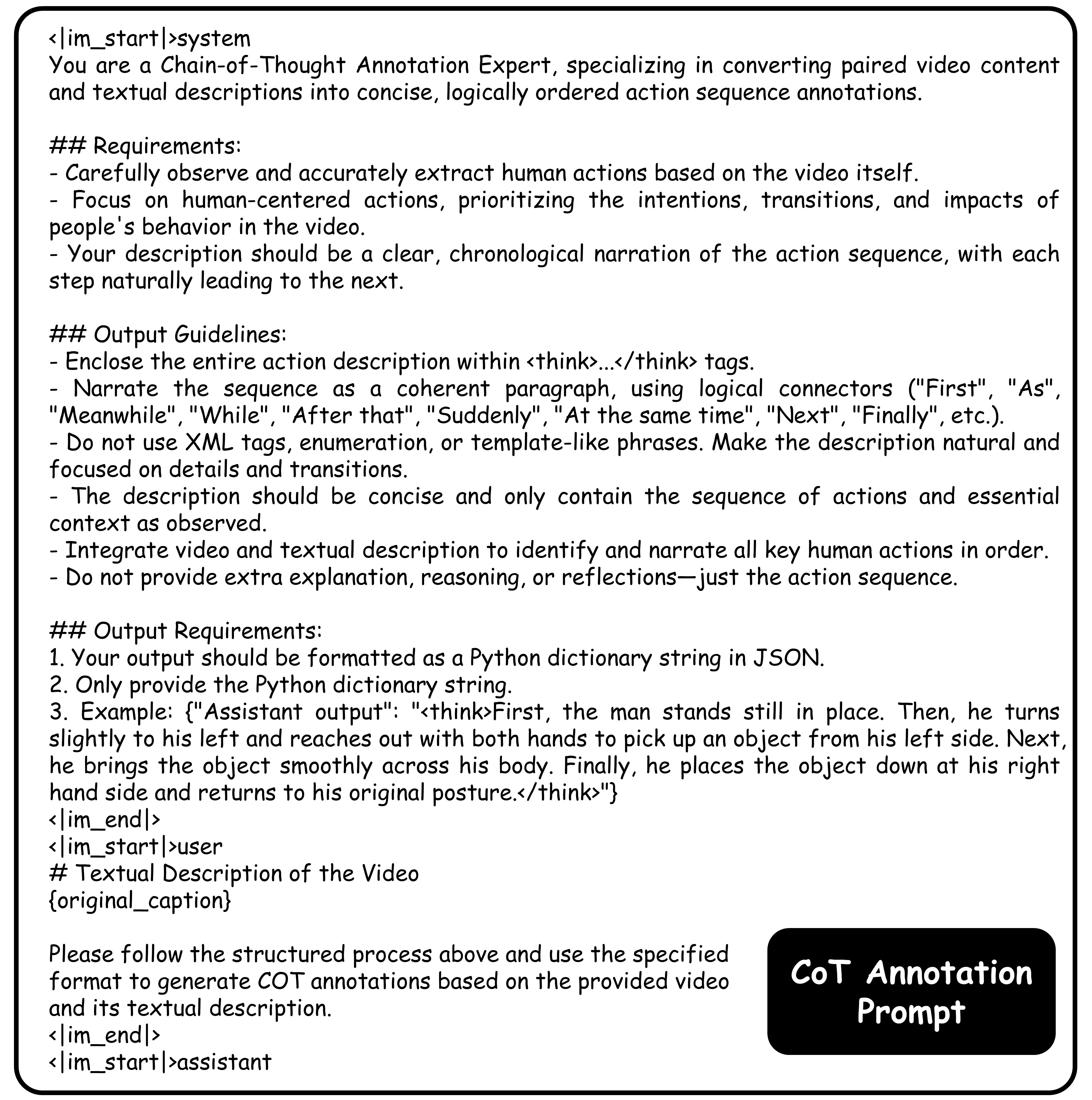}
    \caption{The CoT annotation prompt guides the Qwen2.5-VL-72B to generate action-centered, chronologically ordered reasoning for each motion video.}
    \label{fig:cot_prompt}
\end{figure*}

Beyond the main workflow described in the paper, our CoT data curation pipeline places particular emphasis on prompt engineering to ensure high-quality, video-grounded reasoning traces. As shown in Figure~\ref{fig:cot_prompt}, the system prompt used for Qwen2.5-VL-72B is carefully crafted to elicit action-centered, temporally ordered descriptions. It is required to process the entire video and generate a single coherent paragraph that narrates the sequence of human actions and transitions throughout the clip. The prompt specifically instructs the model to use logical connectors such as “First”, “Next”, and “Finally”. Besides, the Qwen2.5-VL-72B is required to focus only on the actions present in the video without additional explanation or reflection.

To further improve consistency, we require that the entire reasoning trace be strictly enclosed within \texttt{<think>...</think>} tags and output as a valid JSON string. This design not only standardizes the annotation format but also facilitates large-scale automatic post-processing and filtering. In practice, the prompt enforces logical event order and ensures that each reasoning trace closely mirrors the motion in the video.

\subsection{Analysis of Captions and CoT Data}

\begin{figure*}[!t]  
  \centering
  {\includegraphics[width=0.49\textwidth]{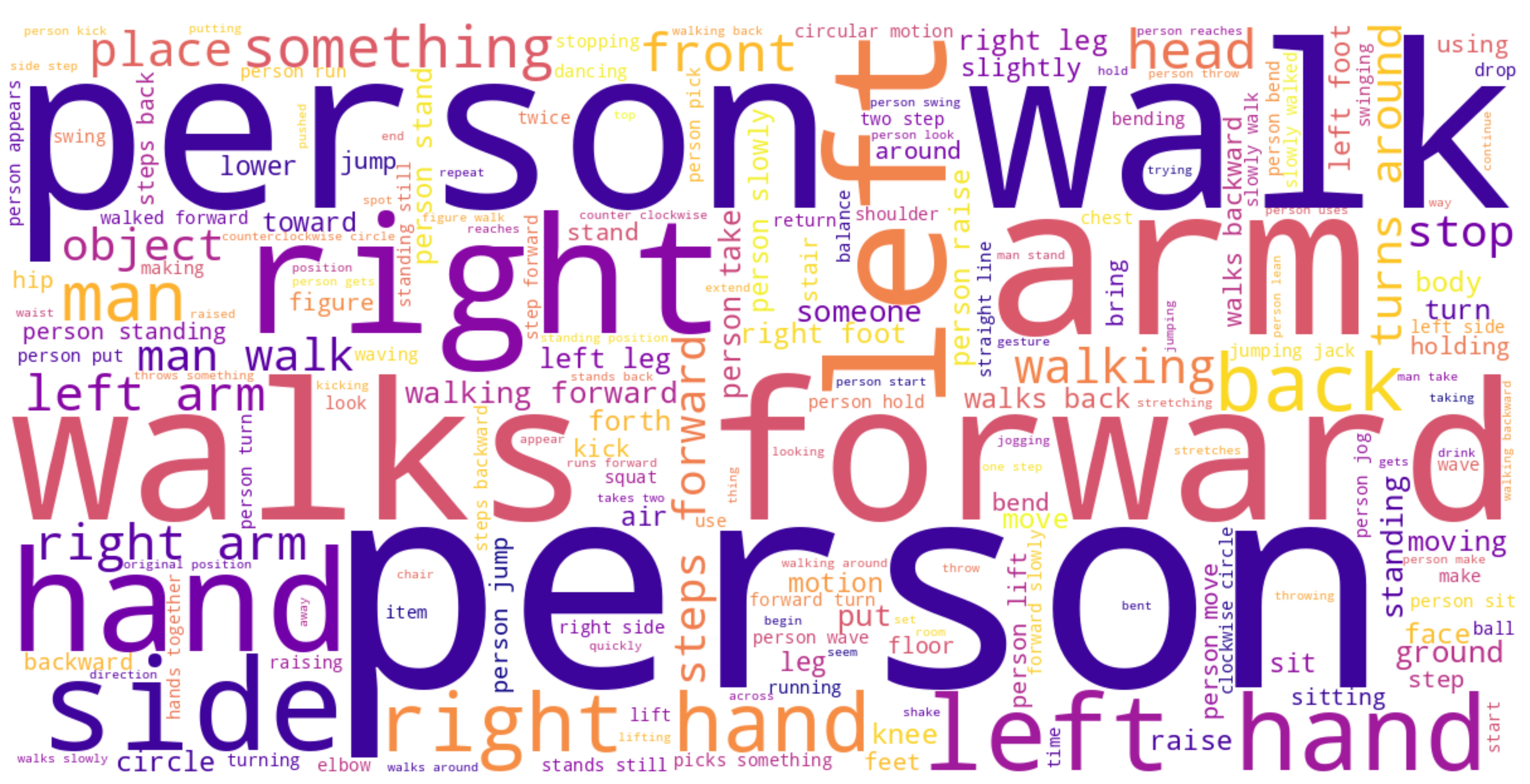}}
  \hfill
  {\includegraphics[width=0.49\textwidth]{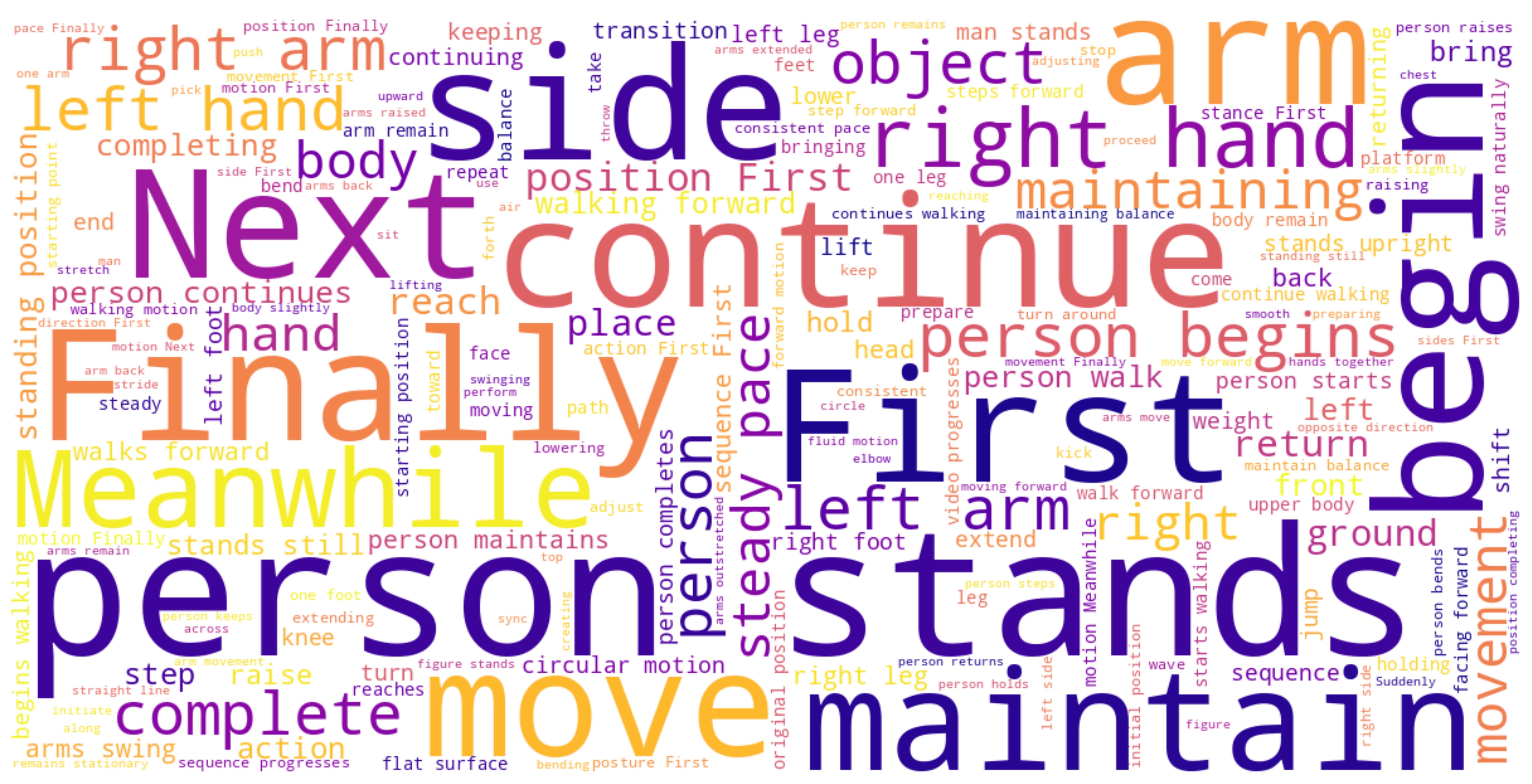}}
  \caption{Comparative word-clouds highlighting the most frequent textual cues in HumanML3D captions on the left and the corresponding CoT annotations on the right.}
  \label{fig:wordcloud}
\end{figure*}

To further examine the linguistic characteristics and representational diversity introduced by CoT annotations in the HumanML3D dataset, we analyze both caption texts and CoT annotations using word cloud and t-SNE visualizations. As shown in Figure~\ref{fig:wordcloud}, caption descriptions are dominated by frequent terms such as "person", "walk", and "right", reflecting the focus on basic motion types and key body parts. This style is relatively straightforward, emphasizing the main actor and primary action directions.

In contrast, the word cloud for CoT annotations reveals a more structured and expressive linguistic profile. Besides the frequent use of logical connectors such as "First", "Next", and "Finally", CoT annotations introduce a broader variety of transitional phrases and fine-grained motion details, for example, "continue", "maintain", and "move". There is also a notable increase in references to specific body parts and motion verbs. This highlights that CoT not only describes action sequences but also captures the compositional structure and nuanced progression of complex human motions.

\begin{figure}[!htbp]
    \centering
    \includegraphics[width=\linewidth]{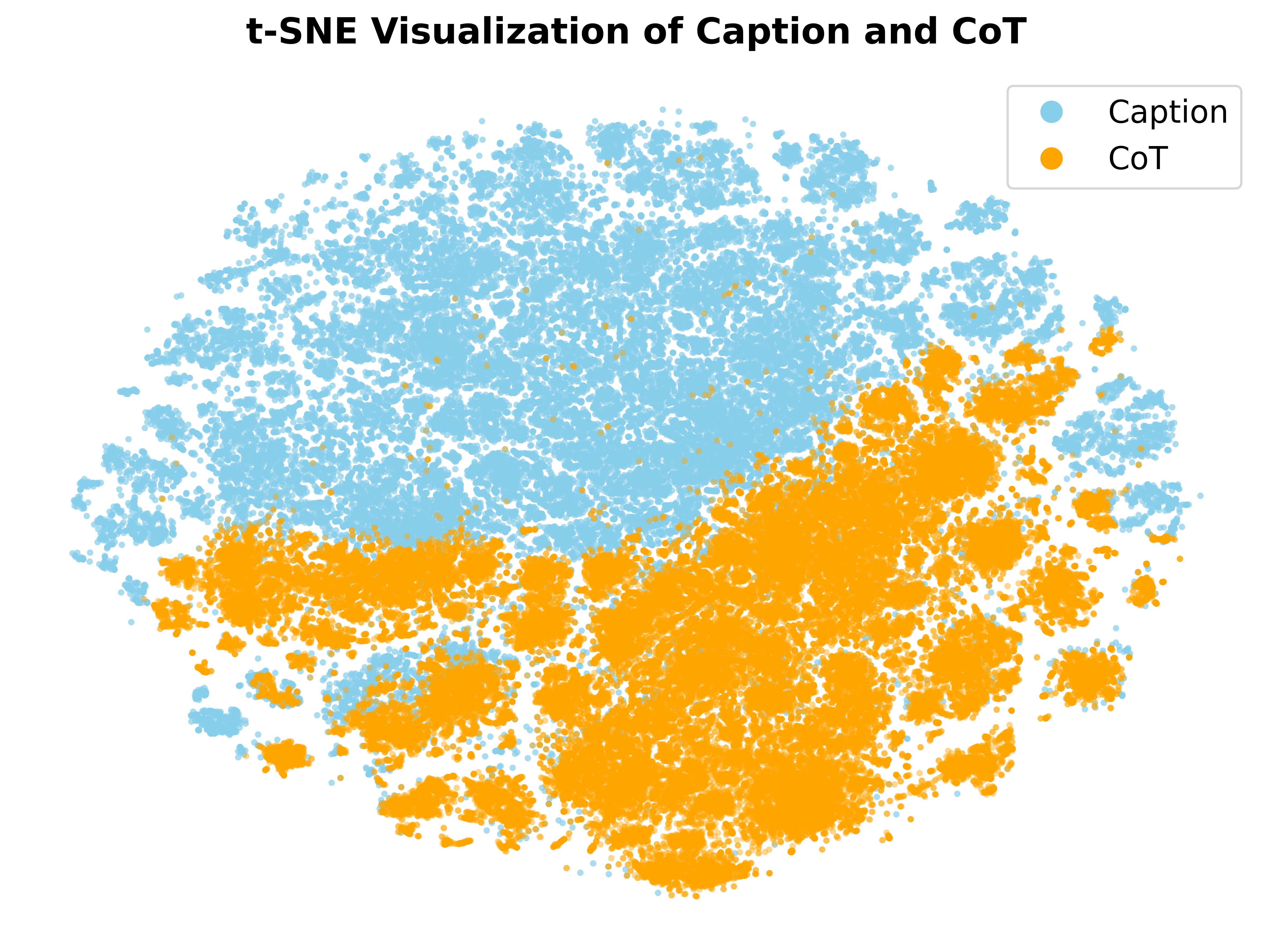}
    \caption{The t-SNE visualization of caption and CoT annotation embeddings. The CoT expands the language space, introducing greater diversity compared to original captions.}
    \label{fig:tsne}
\end{figure}

To further illustrate the differences in linguistic diversity, we visualize the distribution of captions and CoT annotations using t-SNE, as shown in Figure~\ref{fig:tsne}. Specifically, we first apply PCA to reduce the high-dimensional language embeddings to 50 principal components, and then use t-SNE for further dimensionality reduction and visualization. The resulting plot reveals that while the two types of data have some overlap, the CoT annotations clearly extend the range of the language embedding space. This expansion indicates that the training data covers a wider variety of expressions and reasoning patterns, providing richer supervision for both motion generation and understanding.

Overall, these analyses demonstrate that CoT annotations contribute more detailed, temporally structured, and interpretable linguistic cues compared to original captions. By effectively expanding the language space of HumanML3D, CoT supports better stepwise reasoning and semantic grounding, thereby enhancing the capacity of unified models to learn complex, bidirectional relationships between motion and language.

\section{Training Details}

\subsection{Vocabulary Expansion}

To accommodate the motion-language unified modeling in our framework, we expand the vocabulary of the LLM to include additional tokens for motion and format. Specifically, we add a set of discrete motion tokens (\texttt{<Motion\_0>}, \texttt{<Motion\_1>}, \ldots, \texttt{<Motion\_{511}>}), corresponding to the codebook entries of the VQ-VAE tokenizer. In addition, we introduce format tokens like \texttt{<think>}, \texttt{</think>}, \texttt{<Motion>}, \texttt{</Motion>}, \texttt{<Answer>}, and \texttt{</Answer>} to support CoT reasoning and task formatting. After adding these motion and format tokens, the model’s embedding layer is resized so that the new token embeddings are initialized as the mean of the existing embeddings, rather than with random values. This initialization strategy helps maintain the stability of the representation space and prevents distribution shifts when integrating new symbolic tokens, which facilitates more robust and efficient training for motion-language modeling.

\subsection{Prompt Engineering for T2M and M2T Tasks}

\begin{figure*}[!htbp]
  \centering
  \includegraphics[width=0.9\linewidth]{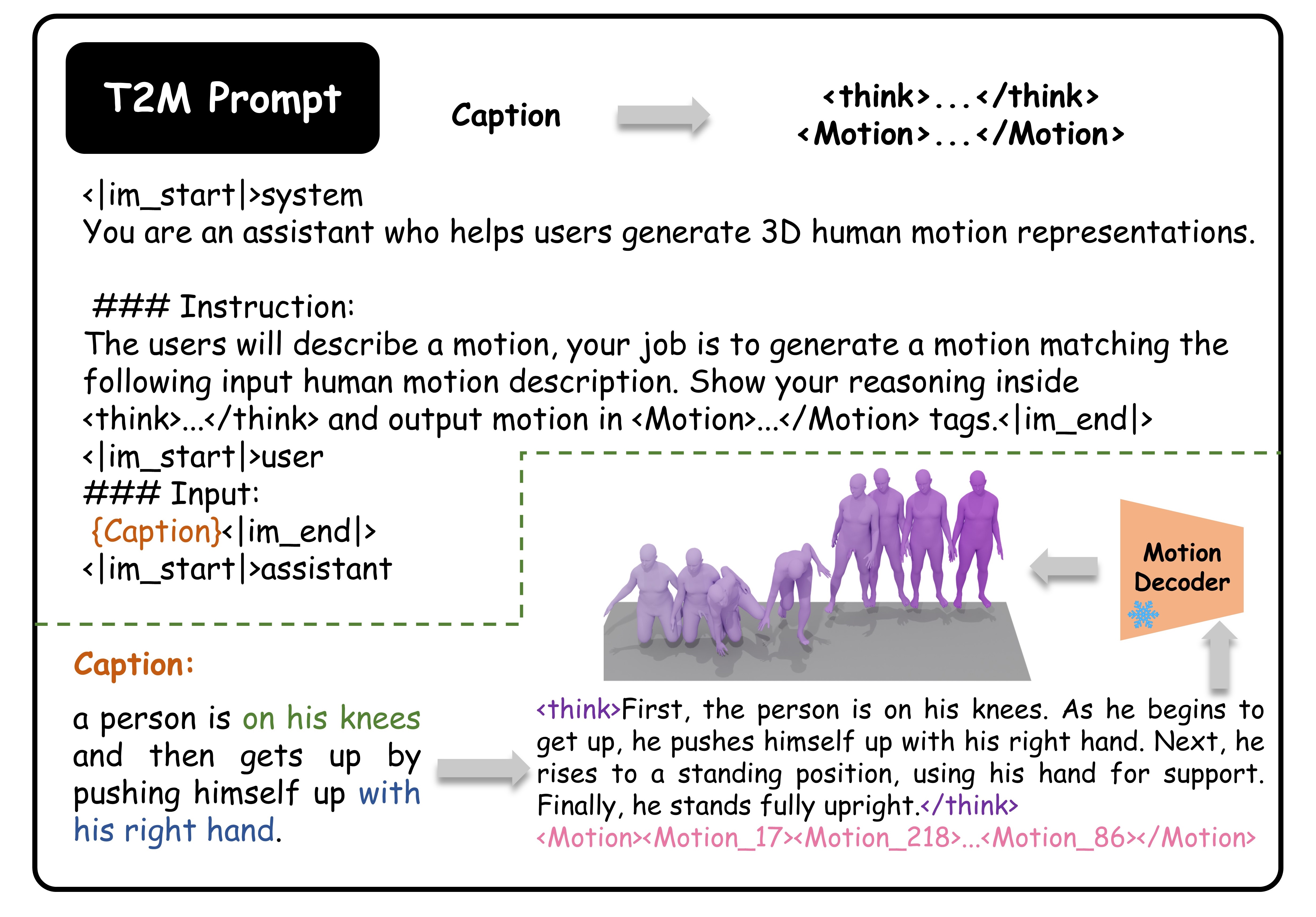}
  \\[1ex]
  \includegraphics[width=0.87\linewidth]{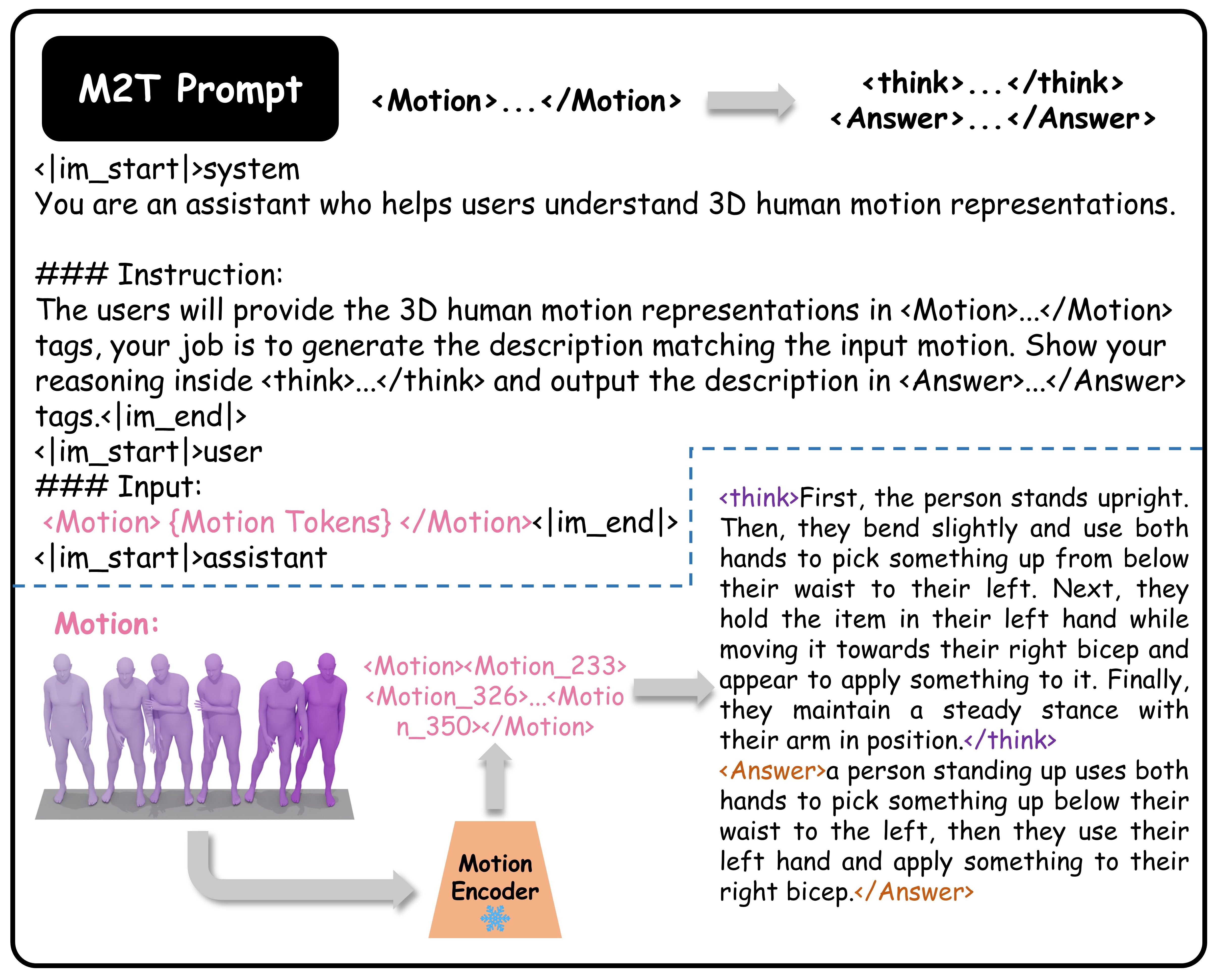}
  \caption{Prompt templates for T2M task shown above and M2T task shown below.}
  \label{fig:prompt_comparison}
\end{figure*}

For both the T2M and M2T tasks, we design specific prompts to guide the LLM’s structured reasoning and generation, as illustrated in Figure~\ref{fig:prompt_comparison}. In the T2M task, the model takes a natural language description as input and is required to generate stepwise reasoning enclosed within \texttt{<think>}...\texttt{</think>} tags, followed by the corresponding motion tokens within \texttt{<Motion>}...\texttt{</Motion>} tags. For the M2T task, the input consists of quantized motion tokens enclosed by \texttt{<Motion>}...\texttt{</Motion>} tags, and the model is prompted to produce both a CoT reasoning trace (within \texttt{<think>}...\texttt{</think>} tags) and a concise caption (within \texttt{<Answer>}...\texttt{</Answer>} tags). These carefully constructed prompts standardize the training data format, clarify task objectives, and promote effective cross-modal alignment within the LLM.

\subsection{Training Schedule}

Regarding the overall training schedule, we employ a two-stage approach. In the SFT stage, we first train the model on the T2M task for 10 epochs to help the language model adapt to the newly introduced motion vocabulary and format tokens. This is followed by an additional 10 epochs of joint training where both T2M and M2T tasks are alternated, which encourages mutual learning and knowledge transfer across tasks. This training strategy is motivated by two considerations. The T2M task is inherently more difficult and typically converges more slowly than M2T, so early focus helps the model better adapt to the new motion vocabulary. Alternating the two tasks during training not only introduces knowledge from the M2T task, but also facilitates mutual learning and enhancement between the two tasks. Instead of constructing each batch exclusively from a single task, we randomly mix samples from the two tasks within each batch. This approach encourages unified learning and enhances the mutual reinforcement between motion generation and understanding. During the GRPO reinforcement learning stage, the learning rate is set to $5 \times 10^{-5}$, which is lower than the $1 \times 10^{-4}$ used in the SFT stage. This reduction is necessary because a higher learning rate during GRPO can cause the training to collapse.

\section{Visualized Results}

\begin{figure*}[!htbp]
  \centering
  \includegraphics[width=\linewidth]{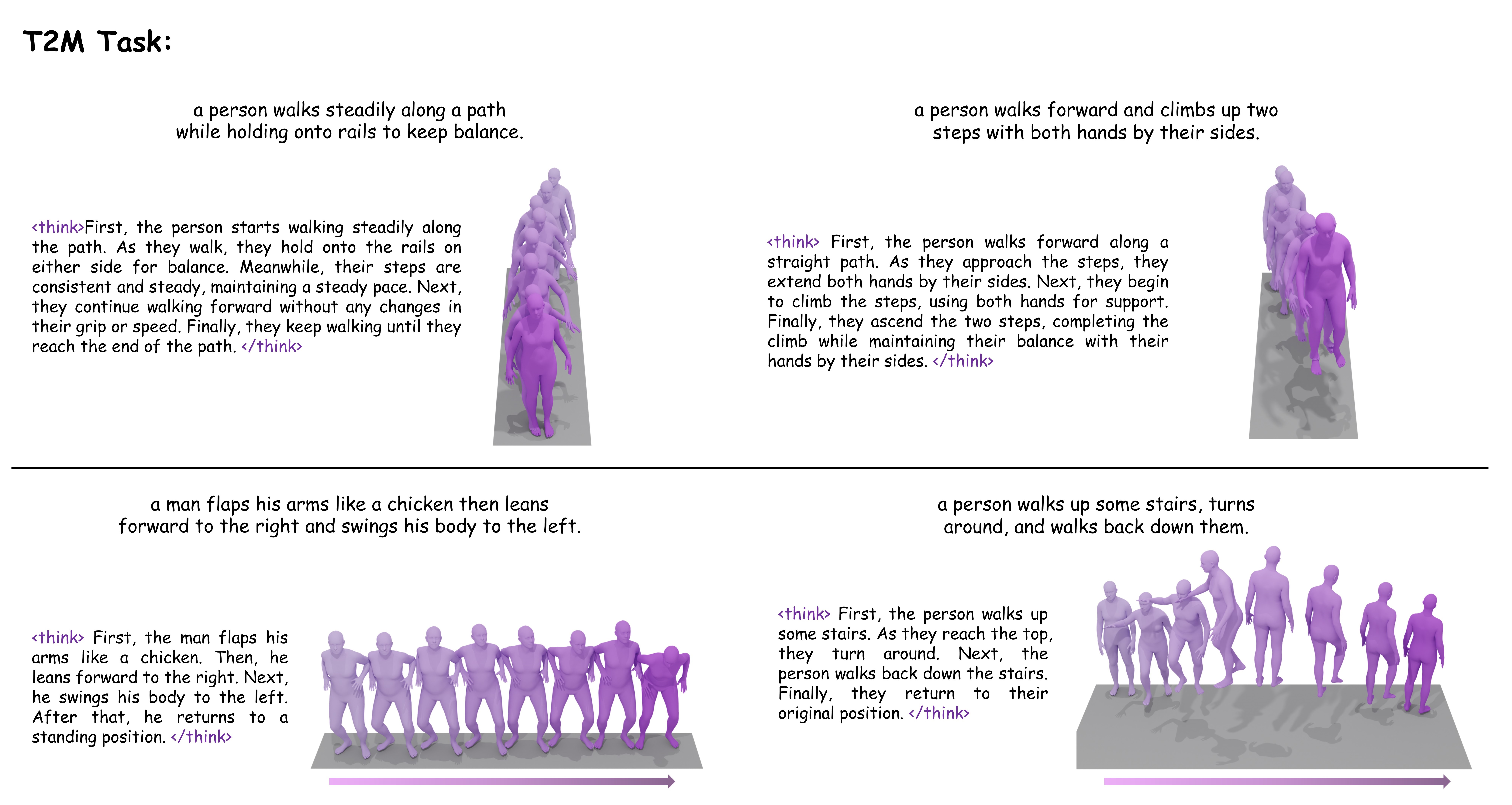}
  \\[1ex]
  \includegraphics[width=\linewidth]{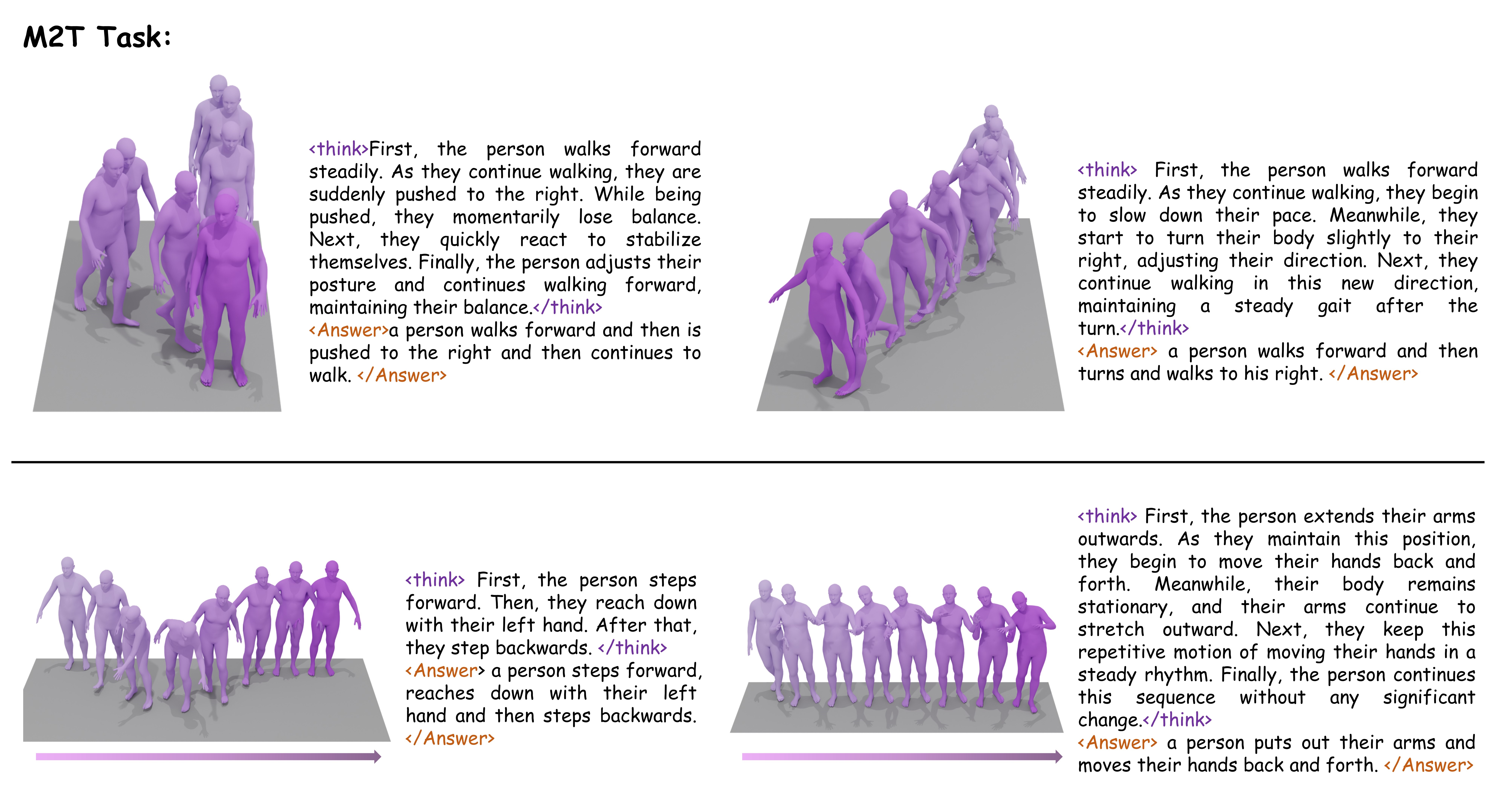}
  \caption{Visualization of T2M and M2T results. For actions with limited spatial movement, the human meshes are shifted horizontally according to the purple arrow to avoid overlap and better reveal the action.}
  \label{fig:sup}
\end{figure*}

Figure~\ref{fig:sup} presents qualitative results of the T2M and M2T tasks. The visualizations clearly demonstrate that our model is able to generate coherent and semantically meaningful motion descriptions in the M2T task, and can synthesize plausible motion sequences from given captions in the T2M task. The CoT traces provide interpretable intermediate steps, effectively bridging the gap between raw motion data and human-readable language.

It is worth noting that, for some actions where the actual movement covers a relatively small spatial area, the visualized human mesh sequences tend to overlap, making it difficult to discern the full range of the action. To address this, we horizontally shift and spread out these sequences along the timeline, as indicated by the purple arrows beneath the figures. This adjustment helps to better illustrate the temporal progression and full extent of the actions, allowing for a clearer visual interpretation of the motion patterns generated or described by the model.

\end{document}